\documentclass[sigconf]{acmart}
\AtBeginDocument{%
  \providecommand\BibTeX{{%
    \normalfont B\kern-0.5em{\scshape i\kern-0.25em b}\kern-0.8em\TeX}}}

\setcopyright{acmcopyright}
\copyrightyear{2021}
\acmYear{2021}
\setcopyright{acmlicensed}

\acmConference[HAI '21]{Proceedings of the 9th International Conference on Human-Agent Interaction}{November 9--11, 2021}{Virtual Event, Japan}
\acmBooktitle{Proceedings of the 9th International Conference on Human-Agent Interaction (HAI '21), November 9--11, 2021, Virtual Event, Japan}
\acmPrice{15.00}
\acmISBN{978-1-4503-8620-3/21/11}
\acmDOI{10.1145/3472307.3484175}

\acmSubmissionID{hai1065}

\begin{document}

\def\code#1{\texttt{#1}}

\title{Say What? \\ 
Collaborative Pop Lyric Generation Using Multitask Transfer Learning}

\author{Naveen Ram}
\email{nram@gatech.edu}
\author{Tanay Gummadi}
\email{tgummadi3@gatech.edu}
\author{Rahul Bhethanabotla}
\email{rmb8@gatech.edu}
\author{Richard J. Savery}
\email{rsavery3@gatech.edu}
\author{Gil Weinberg}
\email{gilw@gatech.edu}
\affiliation{%
    \institution{Georgia Tech Center for Music Technology}
    \streetaddress{840 McMillan Street NW}
    \city{Atlanta}
    \state{Georgia}
    \country{USA}
    \postcode{30318}
}

\renewcommand{\shortauthors}{Ram, Gummadi, Bhethanabotla, et al.}

\begin{abstract}
  Lyric generation is a popular sub-field of natural language generation that has seen growth in recent years. Pop lyrics are of unique interest due to the genre's unique style and content, in addition to the high level of collaboration that goes on behind the scenes in the professional pop songwriting process. In this paper, we present a collaborative line-level lyric generation system that utilizes transfer-learning via the T5 transformer model, which, till date, has not been used to generate pop lyrics. By working and communicating directly with professional songwriters, we develop a model that is able to learn lyrical and stylistic tasks like rhyming, matching line beat requirements, and ending lines with specific target words. Our approach compares favorably to existing methods for multiple datasets and yields positive results from our online studies and interviews with industry songwriters.

\end{abstract}

\begin{CCSXML}
<ccs2012>
   <concept>
       <concept_id>10003120.10003130</concept_id>
       <concept_desc>Human-centered computing~Collaborative and social computing</concept_desc>
       <concept_significance>500</concept_significance>
       </concept>
   <concept>
       <concept_id>10010405.10010469.10010475</concept_id>
       <concept_desc>Applied computing~Sound and music computing</concept_desc>
       <concept_significance>500</concept_significance>
       </concept>
   <concept>
       <concept_id>10010147.10010178.10010179</concept_id>
       <concept_desc>Computing methodologies~Natural language processing</concept_desc>
       <concept_significance>500</concept_significance>
       </concept>
   <concept>
       <concept_id>10010147.10010257.10010258.10010262.10010277</concept_id>
       <concept_desc>Computing methodologies~Transfer learning</concept_desc>
       <concept_significance>500</concept_significance>
       </concept>
 </ccs2012>
\end{CCSXML}

\ccsdesc[500]{Human-centered computing~Collaborative and social computing}
\ccsdesc[500]{Applied computing~Sound and music computing}
\ccsdesc[500]{Computing methodologies~Natural language processing}
\ccsdesc[500]{Computing methodologies~Transfer learning}

\keywords{lyric generation, collaborative AI, transfer learning, transformers, natural language generation, pop music}


\maketitle

\section{Introduction}
Pop music is a genre developed over the last $80$ years, fusing disparate influences into a genre with its own unique lyrical characteristics, based on the goal of mass appeal \cite{frith_straw_street_2001}. Pop lyrics are often created by professional pop songwriting teams collaborating in groups to workshop individual lines until the outcome is as universal, engaging, and ``catchy" as possible \cite{bennett2012constraint}. This environment presents a unique challenge for artificial intelligence based lyric generation, requiring a system which can match the lyrical style and content of a verse in-progress while incorporating constraints given by multiple collaborators. Collaborative text generation models have been explored before, with many providing avenues for users to enrich the generated text with extra examples and seed text \cite{10.1145/2939672.2939679}. More specifically, lyric generation has been explored, with many models replicating whole verses and song fragments in more lyrically dense styles like hip hop \cite{nikolov2020conditional}, but these systems have not yet been adapted to the unique constraints of the pop genre and the use cases of collaborative pop songwriters. 

In this paper we present a collaborative line-level lyric generation system to aid pop songwriters in the rapid generation and refinement of memorable verses. Our system relies on transfer learning, utilizing the T5 transformer to generate lines given a variety of constraints and stylistic settings. We worked with a group of professional songwriters to identify core functionalities and constraints that our model would need to address to participate in their collaborative work setting. This setting is unique within the field of lyric generation and collaboration, as it necessitates the repeated generation of lines with different levels of contexts, meter constraints, and rhyme patterns. A full demo of the system is available online.\footnote{https://tinyurl.com/yywsy456}

By training a model to assist professional song writers and evaluating it against genuine pop lyrics, we hope to provide insight into how state-of-the-art models can be leveraged to accommodate the dynamic demands of human collaborators, how creative AI can be built and customized for interaction, and how transfer learning can be used to apply broader knowledge in fields with limited data. This paper contains a technical overview of our generation system, as well as an analysis of the major constraints of pop lyric generalization. We analyze the data collection, augmentation, and modeling techniques needed to train a dynamic model to address these constraints, and evaluate our models ability to aid songwriters in their creative process and generate lines which match the style and content of the pop genre.

\section{Related Work}
Artificial or model-based lyric generation is a field that intersects several academic areas such as linguistics, musicology, computational creativity, deep learning for natural language processing, just to name a few. At a fundamental level, our task is that natural language generation, a diverse field which has advanced swiftly in recent years with the advent of recurrent neural networks (RNNs) \cite{vinyals2015show,jozefowicz2016exploring}, general adverserial networks (GANs) \cite{guo2018long}, and long short-term memory (LSTMs) \cite{zhang2017adversarial}. The Transformer model introduced recently \cite{vaswani2017attention} is highly successful at natural language tasks when applied through a variety of approaches including auto-encoder models, BERT \cite{devlin2018bert}, auto-regressive models, GPT-2 \cite{radford2019language}, sequence-to-sequence models, and BART \cite{lewis2019bart}.

Despite this broad scope of model architectures and use cases, we found that there were no existing models which applied natural language processing techniques to line-level pop lyric generation. The most relevant tasks which have been thoroughly researched are rap verse generation \cite{savery2020shimon} and poetry generation. In the field of hip hop lyric generation, rhyme is often a focus. One group finds rhyming lines by querying for rhyming words across a vast database of rap song lyrics and picking the most semantically similar line that contains said rhyming word \cite{10.1145/2939672.2939679}. Their approach achieves high rhyme score (or density) results but does not actually generate new lyrics; instead, they find the most suitable next line from their database. Another approach uses a probabilistic model approach that mimics the Blocks Substitution Matrix (BLOSUM) protein homology alignment algorithms to detect rhymes in rap music \cite{hirjeebrown}. While yielding a high rhyme density performance, this method required the explicit creation of a dataset that labeled pairs of rhyming words. Previous approaches, including the ones listed above, make use of either phonemicization libraries like eSpeak \footnote{http://espeak.sourceforge.net/} or Festival \footnote{https://www.cstr.ed.ac.uk/projects/festival/} or self-made linguistics rules to phonemize words and handle pronunciation of slang and words that don’t exist in common English vocabulary. The generation of lyrics has also been considered from a more example-based approach by addressing the problem of rap lyric generation through a statistical machine translation (SMT) model \cite{wu2013learning}. A unique standout of this method is the lack of a use of a priori linguistic or phoneme constraints. Using transformer models, \cite{nikolov2020conditional} provides a baseline for conditional rap music generation from text as well as augmentation to existing rap lyrics.

While rap lyric systems give us an idea of how to approach rhyme density, they differ greatly from pop lyrics in style, themes, and vocabulary. Poetry is a much broader area which includes more established constraints in meter and syllabic placement while often addressing the sort of themes from which pop lyrics are derived.  LSTMs have been used for quatrain generation in sonnets and replicate Shakespearean meter yet cannot match emotion and readability found in human work \cite{lau2018deep}. Chinese poetry generation has seen significant development from RNNs by introducing a combination of a recurrent context model and a recurrent generation model to grasp thematic context over multiple sentences \cite{zhang2014chinese}. Although most of these approaches satisfy coherence and poetic constraints, there is difficulty in generating longer form poetry with meaning. 

Given that contemporary music does not follow perfect rhyme constraints, there is a need to understand the classification of “slant” or “near” rhymes. Previous work presents a methodology, from inspection of human rhyme data, for determining whether two sequences of phonemes are near rhymes \cite{ghazvininejad2016generating}. We adopt a known modification for the aforementioned approach \cite{riedl2020weird} as the revised algorithm accounts for rhyme when expressed vocally, detracting from the strict text-based approach of the former.

\section{System Overview}

In the following section we present, ``Say What?", a novel transformer-based architecture to aid human songwriters at line-level pop lyric generation. Our core challenge is to use given inputs from a group of users to generate a predicted line, which continues the verse in as human-like way as possible by extending the content, lyrical style, and genre conventions of the input verse. Using feedback we received from a group of songwriters, we identify three additional tasks for the project:

\begin{enumerate}

\item Outputs should rhyme

\item Outputs should reflect specified syllable constraints

\item Users should have the option to specify ending words

\end{enumerate}

\subsection{Dataset Collection}

Data collection proved challenging as there is no pre-defined list of pop artists that can be used to outline a dataset. Thus, we approximate the ``pop" genre by filtering artists based on their success or popularity, taking into account that the genre does not necessarily encompass every song that becomes popular. We quantify popularity through the Billboard Hot 100 chart, a standard medium for tracking sales and streams of songs played in the United States. Artists for all weeks from $1958$ - $2017$ were collected and sorted by amount of appearances on chart. The top $25$ songs from each of the top $200$ artists were considered, and their respective lyrics were scraped from Genius - a website for lyric aggregation and annotation. While this approach does include some songs that do not quite fit the structure of pop music stylistically, classifying songs through an objective metric is preferred to a subjective judgement regarding the definition of ``pop".

\subsection{Pre-processing}

After scraping the data, we performed several filtering operations. We first removed songs that were not written in English to ensure consistency with the rhyming pronunciation format we used. To filter out choruses and refrains which are shorter and more repetitive, we deleted consecutive lines with a similarity of over $0.70$ as determined by the python \code{difflib SequenceMatcher} package, short verses (less than $50$ characters long), and verses with less than $6$ lines. Shorter verses were filtered out at a cutoff of $6$ lines in order to be able to generate at least one usable 1-5 line training example from each verse. As pop songs often have repetitive elements, with the same lyrics occurring in successive verses or repeated choruses, we split the train and test sets by song to ensure all the training examples generated from one song would go to the same set. In this manner we avoid evaluating the model on lyrics it may have seen before.

% We generate rhyming dictionaries from both the vocabulary of our dataset and the top $20,000$ most commonly used words in the English language using the \codeword{wordfreq} python module.

\subsection{Rhyme and Phoneme Mappings}
We use the International Phonetic Alphabet (IPA) phoneme representation for word pronunciation, converting each word into its IPA format (using the \code{festival}\footnote{https://www.cstr.ed.ac.uk/projects/festival/} and \code{espeak} \footnote{http://espeak.sourceforge.net/} modules) and save it in a dictionary cache. This allows for quick phoneme comparison between words, giving us the capability to identify rhymes. To account for regional dialect distinctions and for near rhyme we designate a set of phonemes and ending suffixes that are “near” to each other, allowing us to capture cases of near rhyme. Near rhymes occur when last syllables of two words are nearly identical, for example the words ``doing” and ``ruin” are both considered near rhymes of each other despite the former’s ‘NG’ ending in IPA when phonemized. To determine whether two sequences of phonemes are near rhymes we use the methodology presented by \cite{ghazvininejad2016generating}. We adopt a known modification for the aforementioned approach \cite{riedl2020weird} as the revised algorithm accounts for rhyme when expressed vocally, detracting from the strict text-based approach of the former.

\subsection{Model Inputs}

\begin{figure}[h] % add ! after h
    \includegraphics[width=1\linewidth]{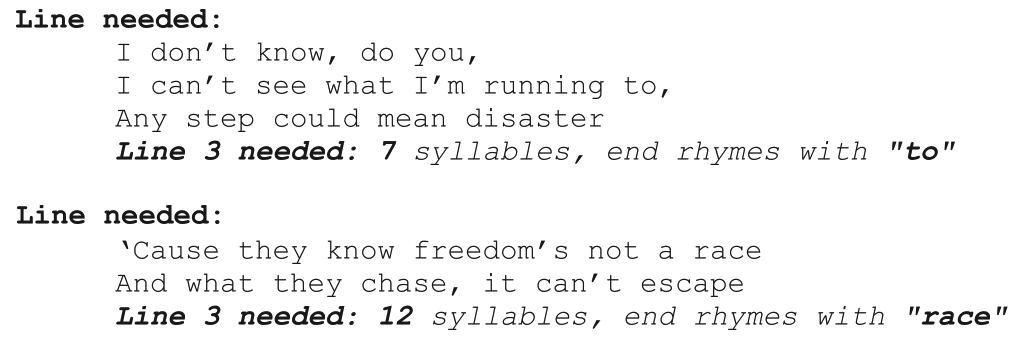}
    \caption{Example use cases given by songwriters}
    \label{fig:songwriterusecase}
\end{figure}

Our training and test datasets were formatted to match the example use cases given to us by the songwriters (see Figure \ref{fig:songwriterusecase}).  

The given examples provided the model with a small number of lines, as input to generate suggestions to continue the verse. Notably, these input lines did not always start at the beginning of a verse, and the model was not expected to finish the verse or provide transitions between verses. To match this functionality, we iterate over verses in our dataset, creating a set of an input lines for each target line and then cutting the length of input lines randomly so that each example has somewhere from $1$ to $4$ input lines and exactly one target line.

\subsection{T5 Transformer Model}

After considering several state of the art transformer models, we decided to use the T5 model published in 2020 by Google \cite{raffel2019exploring}. The T5 model was structured and trained with the intention of leveraging its pre-training knowledge to the widest range of text-to-text tasks possible. The authors opted for an encoder-decoder transformer based design, trained on a variant of the Common Crawl corpus named the Colossal Common Crawl Corpus (or C4 for short). The T5 shows a remarkable ability to tackle many different tasks with one multitask model. It can also perform tasks not commonly approached as text-to-text problems like generating Semantic Textual Similarity (STS) Benchmark scores, generating the numerical response character by character. The T5 is an especially good fit for our task because of the dynamic nature of current and future feature requests by the songwriters. As we analyzed early outputs and writers' requests, the T5 multitask functionality allowed us to incorporate different types of knowledge and accommodate different constraints. We utilize the base size of the pretrained T5 model (220 million parameters), finetuning for 12 thousand additional steps. Our models were trained with a learning rate of 0.003, with a batch size of 128, and a maximum input/output length of 128.

\subsection{Control Model}

\begin{figure}[h] % add ! after h
    \includegraphics[width=1\linewidth]{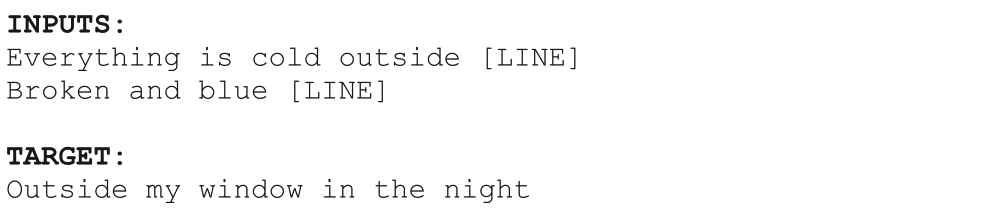}
    \caption{Model input format}
    \label{fig:modelinputs}
\end{figure}

Figure \ref{fig:modelinputs} shows our model inputs. As a final prepossessing step, we separate lines in the input set using \code{"[LINE]"} tokens in order to indicate the divisions in lines to the model. Informed by previous work we add a \code{"finish lines:"} tag to the beginning of the example in preparation for training additional tasks in the future \cite{raffel2019exploring}. Each training example contains 1-4 lines from the verse as the inputs and the following line as the target.

With this data format, we train a model to be used as a baseline for our comparisons, henceforth referred to as the Control model. This model produced promising outputs especially with regards to grammatical, contextual, and stylistic quality of generated lyrics. However, when analyzing results with feedback from the songwriters we found the model fell short in its ability to rhyme and respond to user input. Specifically, the model was not able to take specified syllable count and ending word requirements necessary for pop lyric structure.

\subsection{Rhymes}

\begin{figure}[!ht] % add ! after h
    \includegraphics[width=1\linewidth]{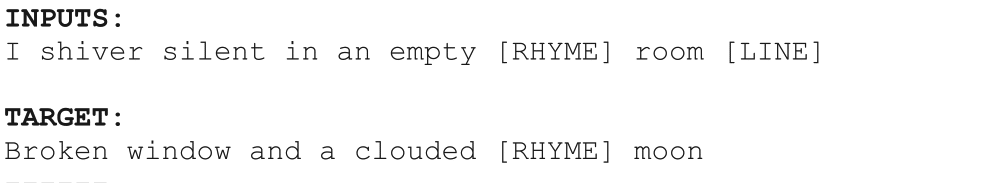}
    \caption{A modified input line including our ``[RHYME]” tag}
    \label{fig:rhymetagexample}
\end{figure}

Less than a quarter of our dataset (exactly $17.3\%$) is composed of rhyming lyrics. In order to generate rhyming outputs, we create a new dataset by filtering out all non-rhyming lines and including a \code{[RHYME]} tag before each line-ending word which rhymed with the last word of the target. An example is shown in Figure \ref{fig:rhymetagexample}. This tag allows the model to associate rhyming words and gain intuition about rhyme.

% In practice, we saw that adding these tags significantly improved our models’ abilities to recognize and generate rhyme. 

\begin{figure}[!ht]
    \includegraphics[width=1\linewidth]{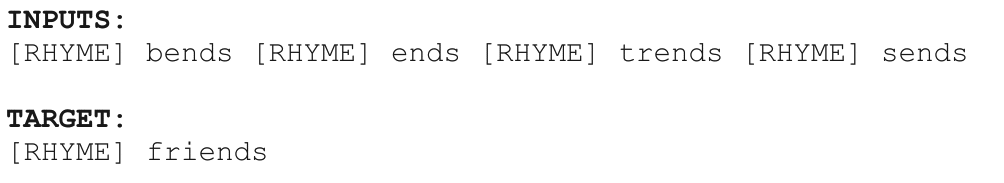} 
    \caption{A rhyme list example}
    \label{fig:rhymelistexample}
\end{figure}

Additionally, we use our pre-built rhyme dictionary to generate a new dataset to be included in a supplementary task we call \textbf{rhyme list}. This task consists of $20,000$ training examples formatted as a list of $5$ rhyming words (separated by \code{[RHYME]} tokens) with one rhyming word as the target (see Figure \ref{fig:rhymelistexample}). These lists provide the model with additional examples of rhyming words to associate.

\subsection{Syllables}

\begin{figure}[!ht] % add ! after h
\centering
    \includegraphics[width=1\linewidth]{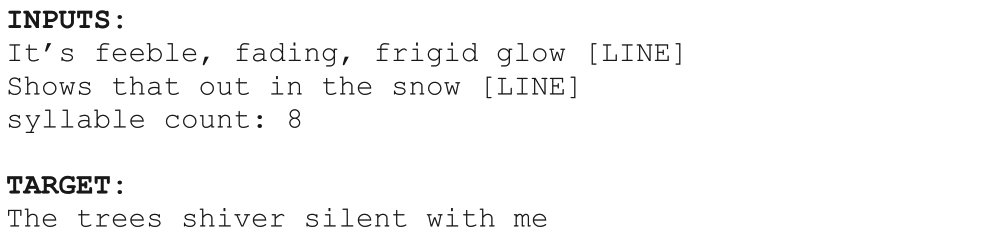}
    \caption{A modified input line with our ``syllable count" tag}
    \label{fig:syllabletagexample}
\end{figure}

To address the songwriter group's request to control for syllable count, we append the syllabic length of the target line as a numerical input preceded by a \code{syllable count:} tag as shown in Figure \ref{fig:syllabletagexample}. This tag enables the T5 architecture to map the given syllable count to the desired syllable count of the line. 

\begin{figure}[!ht]
    \includegraphics[width=1\linewidth]{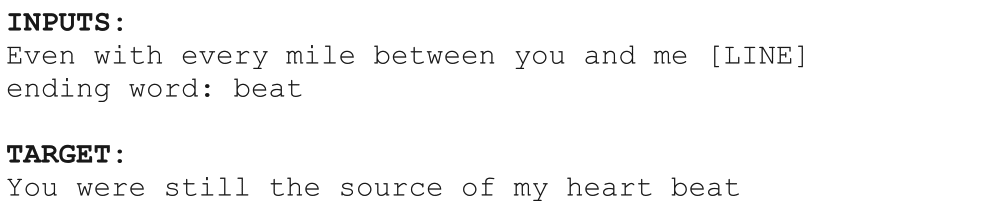}
    \caption{A modified input line with our ``ending word" tag}
    \label{fig:endingwordtagexample}
\end{figure}

\subsection{Ending Words} 

To address the songwriters' third requirement - generating line endings with specific words - a new dataset was created. Mimicking previous methods \cite{kumar2020data}, we append the ending word of the target preceded by the phrase ``\code{ending word:}" to the input as shown in Figure \ref{fig:endingwordtagexample}. The given phrase acts as a label for the expected ending word, helping the model learn to generate lines which end with specific words.

\subsection{Combined Model}
After addressing each problem individually, we created a combined dataset to accommodate different combinations of rhyme, syllable, and end word constraints. To provide the model with consistent inputs, we append the syllable count tag on every example. The trivial case where syllable count is not required is handled by feeding in a count derived from the input lines. Given the limited supply of rhyming data, it is undesirable to overlap rhyming examples with the end word tags, as if a model is given the desired end word, it is not being required to generate a rhyming word on its own. Therefore we split the data into two sets: rhyme and no rhyme.

\begin{figure}[h] % add ! after h
    \centering
    \includegraphics[width=1\linewidth]{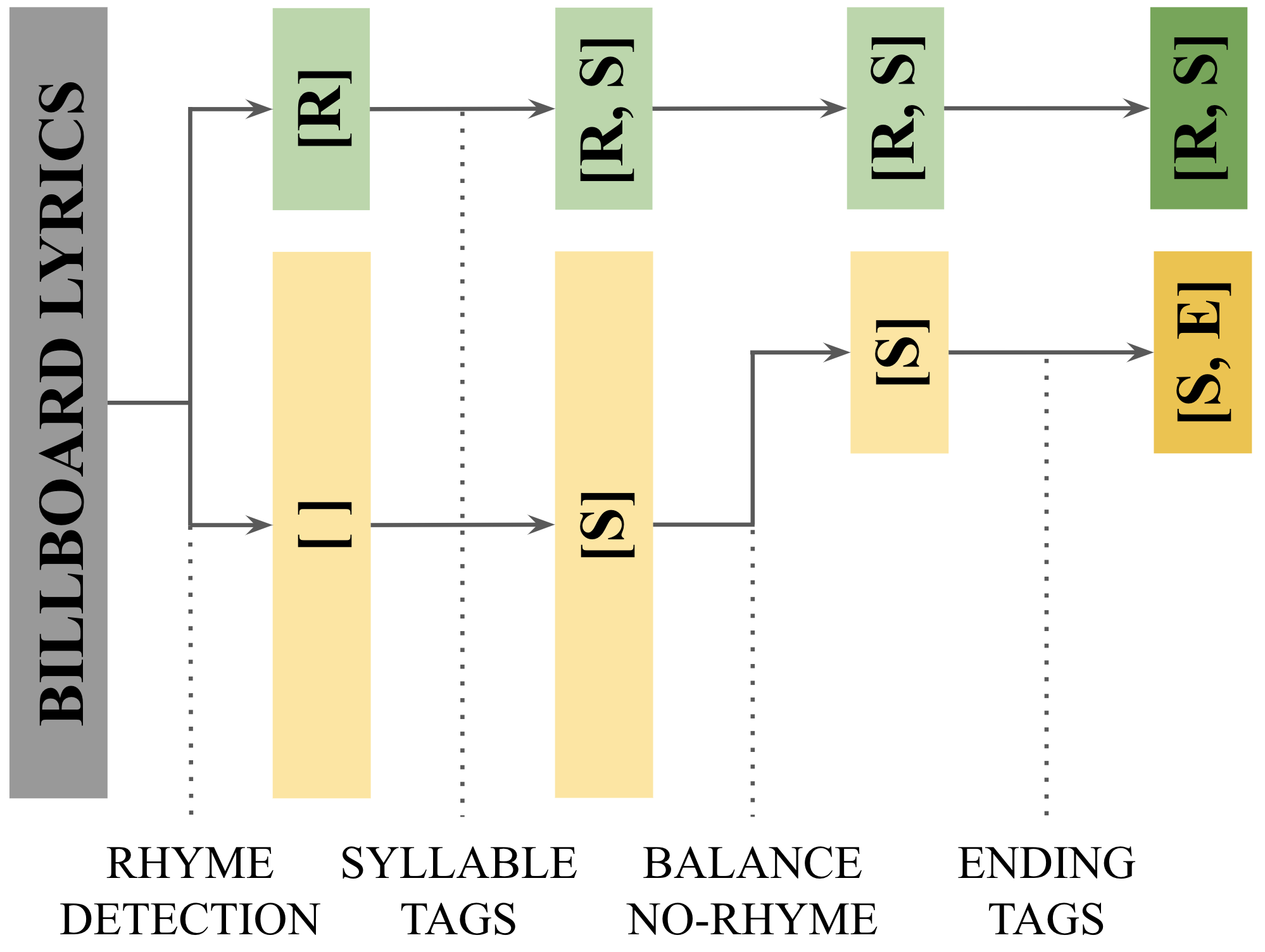}
    \caption{The process for creating the combined dataset including rhyme (R), syllable (S), and ending word (E) tags}
    \label{fig:combinedmodelflowchart}
\end{figure}

 The rhyming set is annotated with \code{[RHYME]} tags and the no rhyme set is formatted with \code{ending word:} specifications. The ending rhyme set is cut to match the size of the rhyming set to balance the data. Subsequently, we orient the two datasets as separate tasks in the T5 model, adding different task tags to the beginning of each example: \code{finish lines rhyme:} or \code{finish lines ending:}. Batches from each task are set to load equally often. This mixture is used in the Combined model. A flowchart of the model combining process is shown in Figure \ref{fig:combinedmodelflowchart}. A final model, referred to as the Combined List model, is trained with an additional third task: the rhyme list task described above in the Rhymes section.

\subsection{Genre Dataset}

Although the main dataset uses the Billboard Hot $100$ to approximate the pop genre, we also created an altered version of the dataset by replacing artists we subjectively deemed unrepresentative of the pop genre with those fitting our intended lyrical style. The resulting dataset shared $67$ of its $200$ artists in common with the regular Billboard dataset, and was pre-processed the same way. Hereon we refer to this new data as the Genre dataset.

\subsection{Demo}

An interactive application was developed and provided for a group of songwriters for evaluation. The application utilized our Combined model, allowing users to provide input lines and constraints. If no constraints were specified, inputs were fed into the rhyme task of the Combined model, using rhyme detection to annotate words which rhymed with the last word given with a \code{[RHYME]} token. The application also allowed users to force a rhyme, feeding in words from the near rhyme dictionary to the ending word task. For each query we found the top $8$ rhymes as sorted by the words' frequency in our dataset. This provided a list of rhyming words common in pop lyrics, rather than a set of random rhyming words. The users could also specify syllable count and end word if they chose. For each query the application generated multiple possible outputs for users to chose from. 

\section{Metrics Results}

In the following section we analyze the performance of each of our models. We are particularly interested to see the extent to which each of our techniques has improved functionality in its target area. The following models listed below in Figure \ref{fig:modeldescriptions} were trained and used for this comparison:

\begin{figure}[h] % add ! after h
    \centering
    \includegraphics[width=1\linewidth]{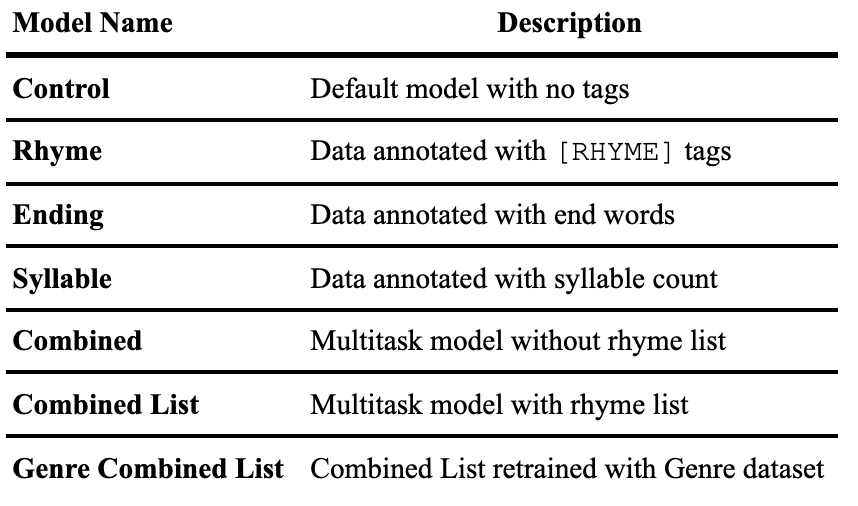}
    \caption{Descriptions of models trained}
    \label{fig:modeldescriptions}
\end{figure}

% \begin{table}[h]
% \begin{center}
%  \begin{tabular}{||l l||} 
%  \hline
%  Model Name & Description \\ [0.5ex] 
%  \hline\hline
%  Control & default model with no tags\\ 
%  \hline
%  Rhyme & data annotated with \codeword{[RHYME]} tags \\
%  \hline
%  Ending & data annotated with end words \\
%  \hline
%  Syllable & data annotated with syllable counts \\
%  \hline
%  Combined & multitask model without rhyme list\\
%  \hline
%  Combined List & multitask model with rhyme list \\ 
%  \hline
%  Genre Combined List & Combined List model retrained \\ \ &  with the Genre Dataset \\ [1ex] 
%  \hline
% \end{tabular}
% \caption{Descriptions of the models tested}
% \label{table:modeldescriptions}
% \end{center}
% \end{table}

To quantify our model’s ability to generate authentic and semantically meaningful lyrics, we test our outputs based on three major categories: grammatical and syntactic sense, lexical similarity to our inputs, and artistic quality of our outputs. Our decision to do so, and the specific metrics we chose were based off previous literature surrounding the assessment of machine-generated text \cite{syntheticliterature,statisticalfeatures}, the musical qualities common to pop music like rhyme \cite{metricaltechniquesrap}, as well as features related to comments made by songwriters we had spoken with before adding syllable count and end word accuracy. We used BLEU score to measure how similar our outputs were in relation to our training data to capture our models' abilities to create lyrics of human-written quality. \cite{papineni2002bleu}.
To measure lexical similarity between our predictions and our targets, we took the difference between the two to ascertain if our inputs were not significantly less (or more) verbose than our inputs; conceivably, we assumed, if the last line of a verse had a wider range of vocabulary than the rest of the verse, it would be a clear indication of machine-generated lyrics. Finally, we evaluate the categories specified by the songwriters; namely, the ability to rhyme, match a syllable requirement, and to end a line with a specifically inputted word.  

\begin{figure}[h] % add ! after h
    \centering
    \includegraphics[width=1\linewidth]{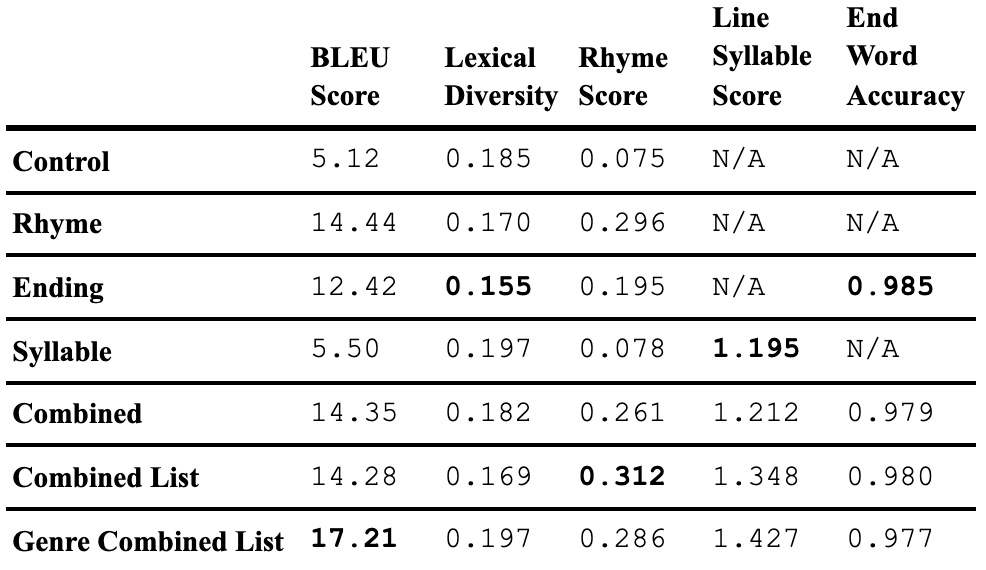}
    \caption{Results across all our models}
    \label{fig:bigresultstable}
\end{figure}

\subsection{BLEU}

\begin{figure}[h] % add ! after h
    \centering
    \includegraphics[width=1\linewidth]{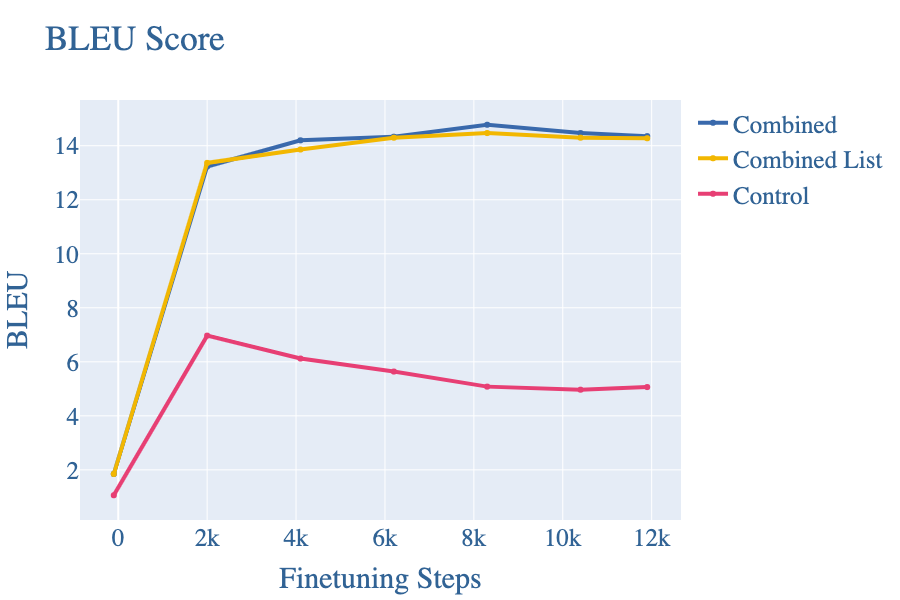}
    \caption{BLEU scores for the Control, Combined, and Combined List models}
    \label{fig:bleuscores}
\end{figure}

As it has been a standard for assessing the quality of machine-generated text and has seen use in a variety of transformer-based text generation tasks, we opt to use BLEU score as an assessment of the similarity of our model outputs’ with the actual lyrics, which are used as targets \cite{bleuscoreusemachinetranslation,bertusesbleu}. Additionally, the use of the BLEU score provides us with a metric to compare the quality of our various models. As seen in Figure \ref{fig:bleuscores}, the Combined and Combined List models achieve similar performance at $14.35$ and $14.28$, respectively. Lyric generation is an open ended task, and we are asking our model to generate a relatively unconstrained line when given a small amount of context. Given such a broad task, achieving a BLEU score of $14$ is impressive despite being much lower than scores achieved by state-of-the-art models in more constrained tasks such as machine translation; for example a transformer based architecture for conditional rap lyric generation achieved a BLEU score of $14.3$ among the best models \cite{nikolov2020conditional}. The Control model performed significantly worse with a score of $5.065$. The key difference here is the size of the datasets. As the combined datasets were filtered to have an even number of rhyme and non-rhyme examples, the respective models were more focused and able to replicate the targets (the original song lyrics) more accurately. This reflects a general principle of transfer learning: that fine-tuning on too much data can be detrimental to a model's performance \cite{ayub2020cognitively}.

\subsection{Lexical Diversity}

% \begin{figure}[h!] % add ! after h
%     \centering
%     \includegraphics[width=8cm]{images/Lexical Diversity.png}
%     \caption{Lexical diversity RMSE scores for all the models}
%     \label{fig:lexicaldiversity}
% \end{figure}

The Lexical Diversity Root Mean Squared Error (RMSE) calculates the RMSE for the Type Token Ratio difference between predictions and targets. Among the different input data/models, all had differences in the range of $0.15$ to $0.2$. The Genre Combined List data had the greatest difference in prediction/targets with an RMSE of $0.197$ and the Ending Word model had the least difference with an RMSE of $0.155$. Over many iterations the lexical diversity RMSE is minimized as the T5 moves from arbitrary generation to matching the style and lexicon of the given lines. It is important to note that given our relatively small input data, the model is able to use the context and style of the inputs to infer the amount of lexical diversity to use in its generated output. % Refer to Figure \ref{fig:lexicaldiversity} for a comparison of lexical diversity rmse across the models. 

\subsection{Rhyme Score}

\begin{figure}[h] % add ! after h
    \centering
    \includegraphics[width=1\linewidth]{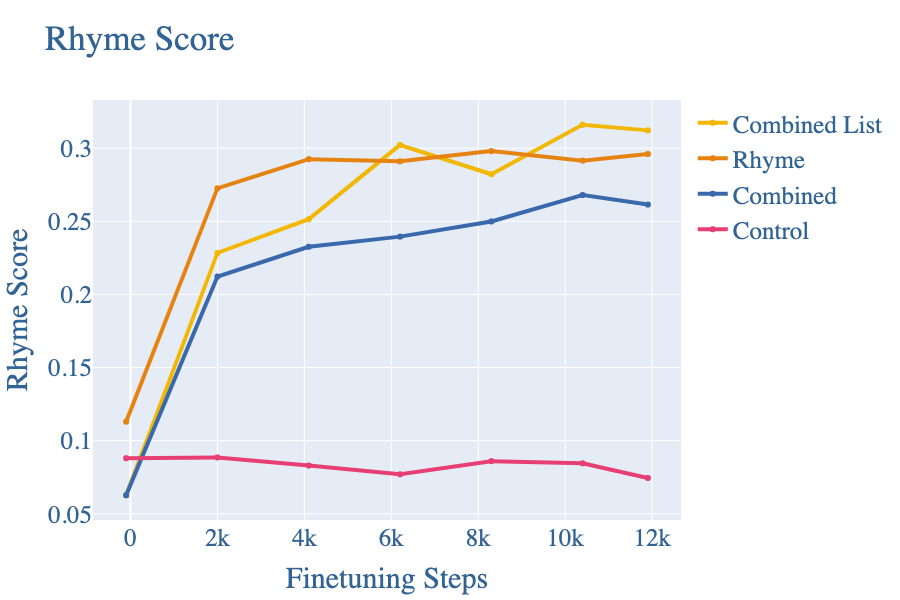}
    \caption{Rhyme scores for the Control, Combined, and Combined List models}
    \label{fig:rhymescores}
\end{figure}

Rhyme Score is a metric to quantify how often our models were exhibiting an end-word rhyme pattern. The metric is calculated as the percentage of the predicted lines which end with a word that rhymes with one or more of the end words of the input lines. Rhyme detection is done following the methods specified in \cite{riedl2020weird} to compare whether two words are near rhymes. The Control model was not able achieve a rhyme score of more than $.0745$ due to the lack of rhyming examples in its training data. When the dataset was filtered and \code{[RHYME]} tags were added, the resulting model (Rhyme) was able to increase in score by $297\%$, achieving a final score of $0.296$. When the ending word examples were added in the creation of the Combined model, the proportion of rhyming inputs decreased to $50\%$, but the model was still able to achieve a high rhyme score of $0.2614$. Finally, the inclusion of the rhyme list task boosted this score by $19.4\%$, achieving our best overall performance for the Rhyme Score metric: $0.3122$ with the Combined List model. The progression is visualized in Figure \ref{fig:rhymescores}.

\subsection{Line Syllable Score}

\begin{figure}[h] % add ! after h
    \centering
    \includegraphics[width=1\linewidth]{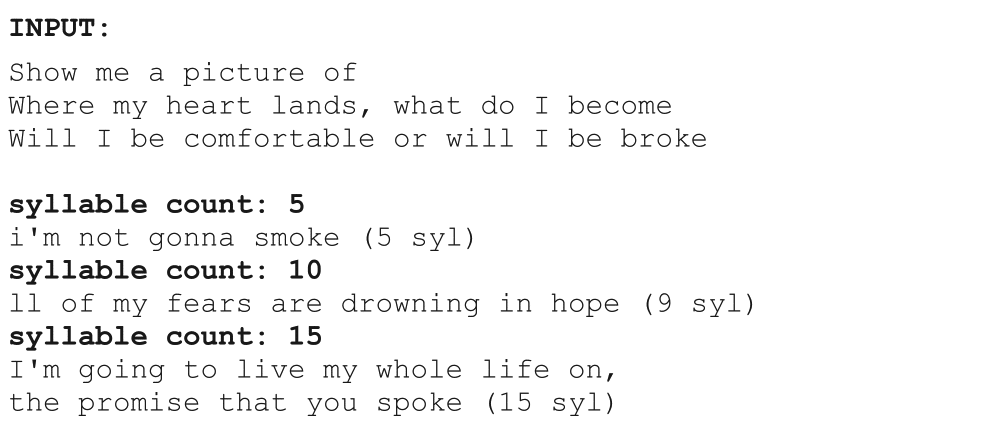}
    \caption{Model outputs when given 5, 10, and 15 syllable count tags for the same input}
    \label{fig:syllableoutputexample}
\end{figure}

The Syllable RMSE calculates the difference in syllable counts between the predictions and targets. Every model exhibited an RMSE below $1.5$ syllables with the standard dataset performing marginally better (lower RMSE) than the Genre data (see Figure \ref{fig:bigresultstable}). When demoing for the songwriters we found that an error of one or two syllables was acceptable because lyrics can easily be adjusted by contracting or expanding common words. Even when we enforced our syllable requirement as a hard constraint, the model was almost always able to generate the correct syllable count within the first $5$ generations. Figure \ref{fig:syllableoutputexample} demonstrates the model's ability to match a broad range of syllable requirements despite unchanging inputs. 

\subsection{End Word Accuracy}

\begin{figure}[h] % add ! after h
    \centering
    \includegraphics[width=\linewidth]{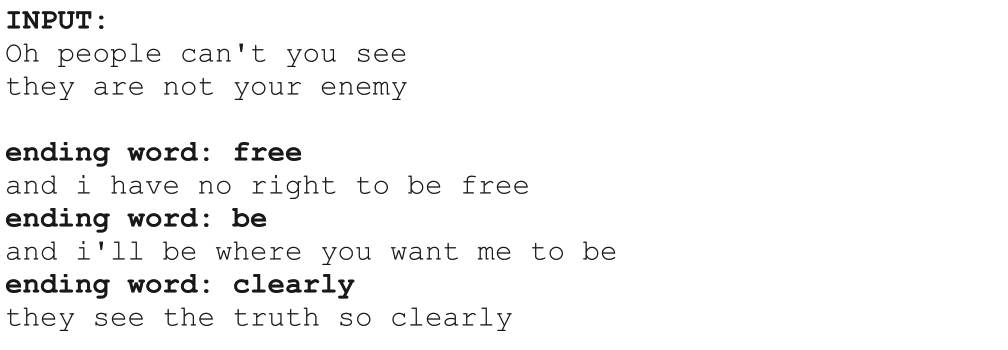}
    \caption{Model outputs when given three different ending word tags for the same input}
    \label{fig:endwordoutputexamples}
\end{figure}

The End Word Accuracy calculates the ratio of ending words matching between each line of predictions and each line of targets. All of the models excelled for our songwriting purposes, displaying accuracies greater than $0.97$ in matching ending words of predictions and targets (see Figure \ref{fig:bigresultstable}). Note in Figure \ref{fig:endwordoutputexamples} that the generated lines integrate the context of the input lines with the ending word, leading each line to a natural conclusion.

\subsection{Genre Dataset}

When compared to the standard models trained with the Billboard dataset, the Genre Combined and Genre Combined List models trained using our hand-picked pop genre dataset performed comparably in all areas except BLEU Score where it outperformed its counterparts by $23.9\%$, achieving a BLEU Score of $17.21$. This increased performance suggests that when given a dataset which conforms more strictly to the lyrical style of the pop genre, our model is able to match the desired targets closely.

\subsection{Model Conclusions}

The best models were the Combined List and Genre Combined List models as they achieve high BLEU scores while also scoring comparably if not better than other models in Lexical Diversity, Rhyme Score, and End Word Accuracy. Both models perform the three tasks the songwriter specified, namely both satisfy rhyming, syllable, and end-word functionality without significantly sacrificing performance in any of those categories when compared to the task specific models tailored for the aforementioned functions. Thus the results show that we were able to successfully combine all three of the functionalities without sacrificing performance in any meaningful way in our metrics.

\section{Human Evaluations}

In addition to analyzing quantitative measures of success, we gathered qualitative assessments of our model and its ability to produce human-like lyrics. We performed an online study using Amazon’s Mechanical Turk (MTurk) as well as conducted interviews with songwriters to receive feedback on the usability and quality of our model. 

\subsection{Mechanical Turk Online Study}

The MTurk study included $67$ participants residing in the United States and India from the ages of $26$ - $75$. Of these $67$ participants, $4$ were eliminated for not passing an attention check. The remaining 63 participants were asked to answer a variety of questions ranking the quality of the following 3 models: Combined List, Genre Combined List, and Control. It is important to note that the participants are not domain experts in songwriting nor potential users of the system, but rather a representation of future listeners who will experience the outputs generated by the system. The study is intended to evaluate the quality of outputs provided and to assess whether generated lyrics are distinguishable from actual lyrics to the common listener, the audience pop music is specifically designed for. 

The participants were first asked to rank outputs based on grammatical fit, stylistic fit, rhyme, and logical coherence. A second set of questions, in the style of a Turing Test, asked participants to predict the real line between a model output and the original (human) ending line of each provided verse. Although a Turing style format is not ideal for a creative task of lyric generation, the test is useful in combination with the aforementioned quantitative metrics. Analysis of the results is discussed below, excluding cases where participants indicated they recognized the provided verse.

\begin{figure}[h]
\includegraphics[width=1\linewidth]{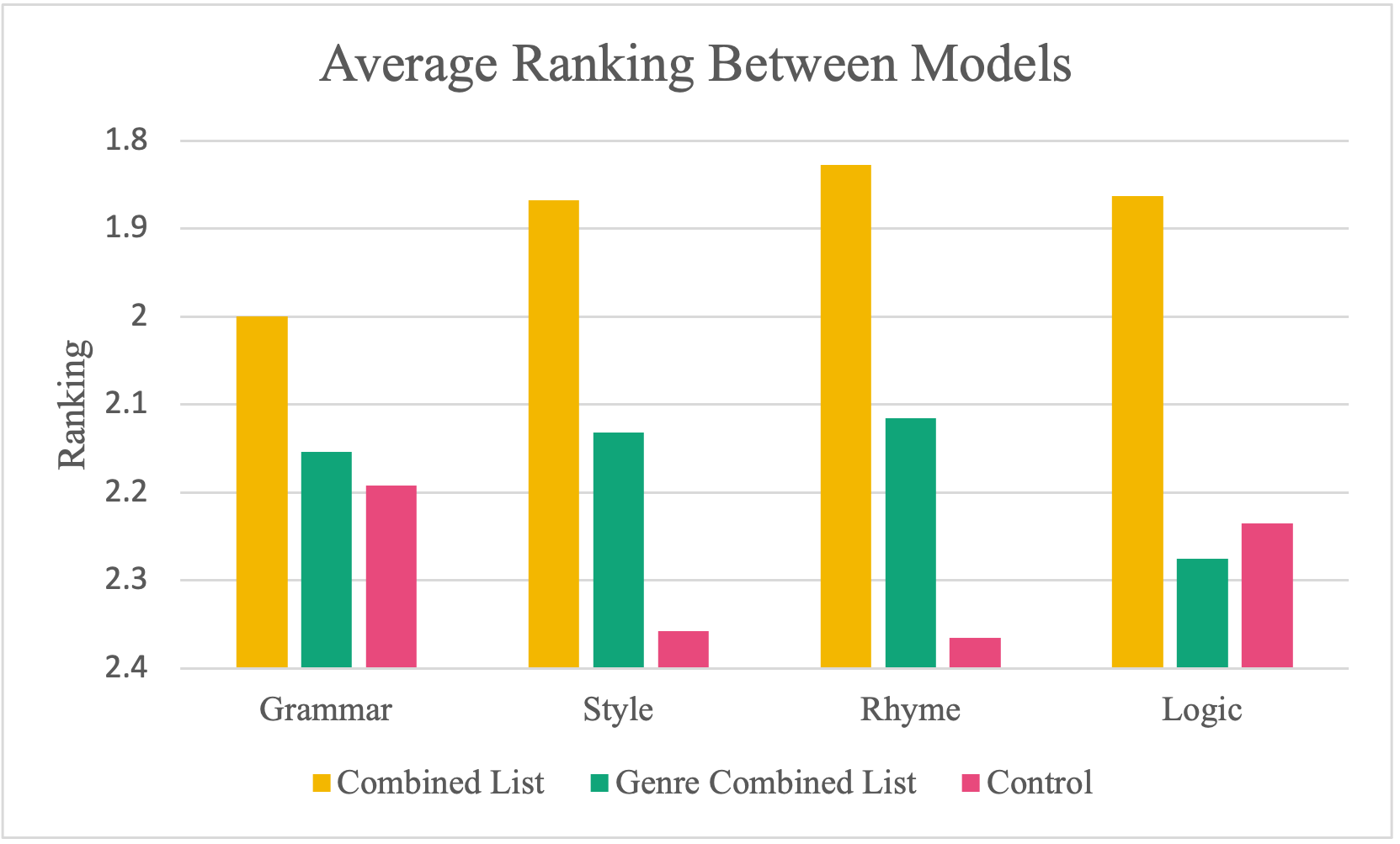}
\caption{Average ranking across grammar, style, rhyme, and logic}
\label{fig:modelrankings}
\end{figure}

For the first set of questions, the Combined List model was ranked the highest for all the categories and the Control model the lowest (refer to Figure \ref{fig:modelrankings}). Conducting an ANOVA between the models exhibits the following p-values for the given categories: grammar ($p = 0.139$), style ($p = 0.002$), rhyme ($p < 0.001$), logic ($p = 0.0126$). There was no statistical difference in the rankings for grammar, but for style, rhyme and logic there were statistically significant differences in the generated outputs. These average rankings (displayed in Figure \ref{fig:modelrankings}) indicate that the lines generated by the Combined List model were superior in these three categories.

\begin{figure}[h]
\includegraphics[width=1\linewidth]{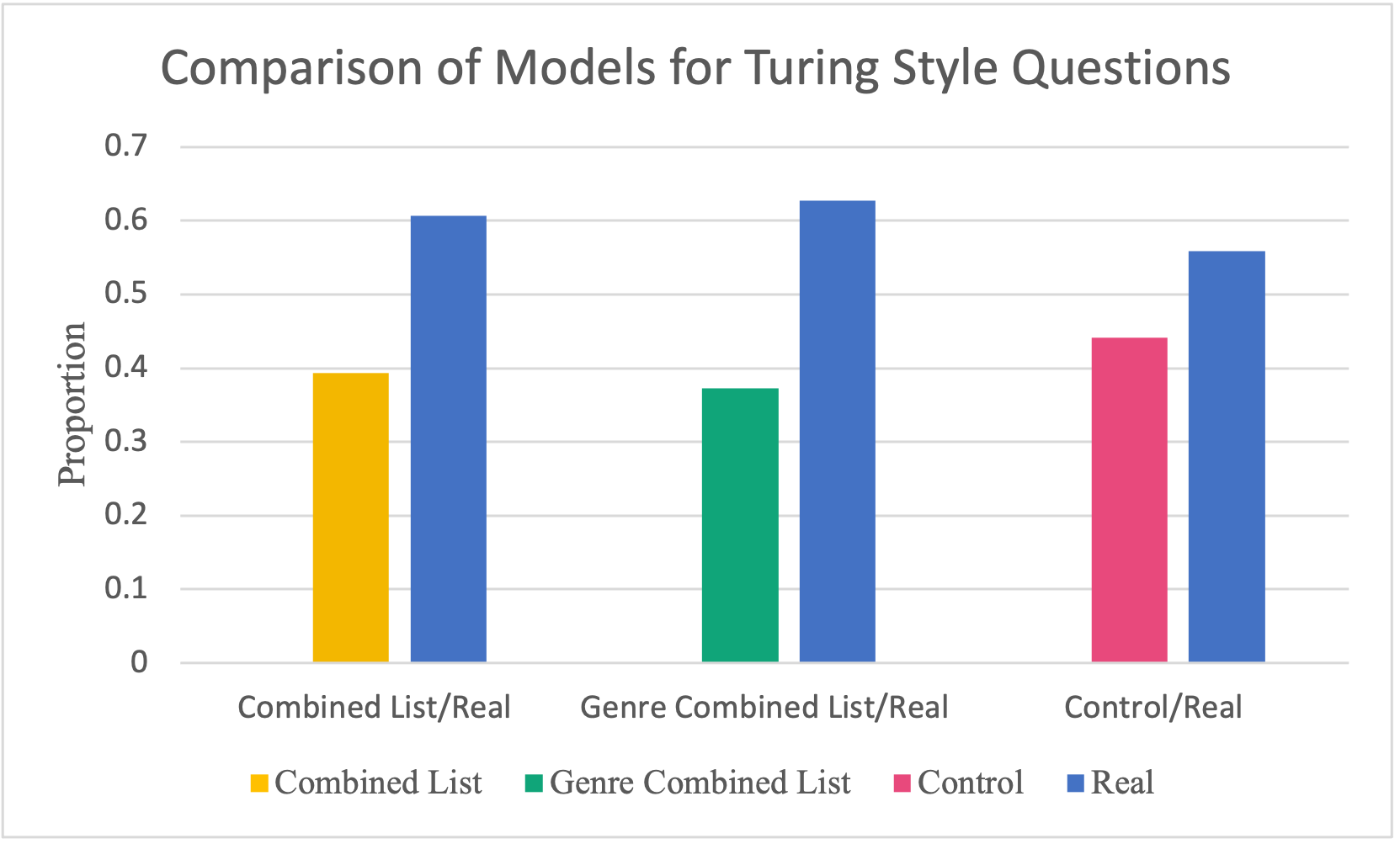}
\caption{Turing Test style question results comparing model outputs to actual lines}
 \label{fig:combinedturing}
\end{figure}

In asking participants to guess the real line, between model outputs and the actual line, the Combined List model and Genre Combined List were both chosen approximately $38\%$ of the time whereas the Control model was chosen $44\%$ of the time (refer to Figure \ref{fig:combinedturing}). Our architecture vastly outperforms previous results accomplished in hip-hop \cite{nikolov2020conditional} in the proportion of times participants were fooled. A pairwise t-test across the models, corrected by the Holm–Bonferroni method, yields p-values of $0.466$, $0.155$, $0.704$ for the Combined List, Genre Combined List, and Control models, respectively, when compared to the real lines. There is no significant difference between the models and the real lines.

\subsection{Songwriter Interviews}
We interviewed three songwriters who had not been involved in the creation of the system aiming, to understand potential use cases for the lyric generator and to gather their thoughts on the system in an informal setting. We opted for longer, informal interviews in order to evaluate the songwriters experience as potential users of the system, a method which has shown past success \cite{savery2020shimon}. This also provided a contrast to the the shorter survey we performed with our MTurk participants, who represented the potential listeners. The participants were professional songwriters working in the US. We allowed each song writer to freely interact with a demo of the system that generated one line, based on a three line input, as well as a five line generation. In their interactions with the system two of the song writers started by instantly adding existing song lyrics to the system, to see how close the system came to the original lyrics. They then gradually changed these lyrics to see the effect, before adding new lyrics written on the spot or other preexisting lyrics.

From these interviews several threads emerged. All three described the potential use of the system as an ``idea bank", that they would try many times for a lyric and filter through the results. They could also all imagine using the system to help further develop ideas and expand their own work. The sophistication in rhymes was well-recognized, and especially the use of near rhymes such as ``mobile" and ``local"; one songwriter noted that the ability to rhyme was ``exceptional" and by the far the best feature of the system. One songwriter commented that the generations often took the form of ``It’s not quite this, but if I change this word it’ll be perfect", describing the system as someone you can ``shoot ideas off" for ``instant feedback". 

Between the three musicians there were several requests and ideas for improvement. A common thread amongst all three was a desire to be able to set the ``mood", ``style", ``genre", or ``tone" of writing from the system. All three also noted a desire to be able to change the rhyme patterns, from only rhyming the last word, to more sophisticated forms. One songwriter mentioned the desire to be able to choose which rhymes lined, such as each second, or more complicated structures such as the blues form. Two of the song writers noted that in its current form the system is ``not very good if you don’t have any ideas", and is a better tool if you come in with ideas to expand on.  

Overall all three writers enjoyed the functionality of the system as something to expand their own work. One songwriter noted that it fulfilled what they saw as the function of computers in their work: to assist and improve their process while not necessarily replacing it. Another songwriter echoed this idea, stating that if they were to use AI in their process they would request for lyrics reflecting their needs, and quickly ignore parts that they don’t like. All three writers requested access after the interview to continue using the system.

\section{Discussion and Conclusion}

We built ``Say What?", a tool for line-level lyric generation in the collaborative atmosphere of pop songwriters. We approached this challenge as a line-level pop lyric generation transfer-learning task. To that end, we trained a transformer architecture to learn stylistic features of pop music by passing in pre-processed data. Multiple models were trained to learn these various tasks and evaluation was performed using pertinent genre metrics including rhyme density, consistency with expected syllable counts, and BLEU score.

The results of our study reveal several takeaways. From a quantitative perspective, the ability of our Combined List models to integrate the functionalities of the three songwriter tasks without significantly sacrificing performance in any of our metrics reflects on the potential of combining tagging-based pre-processing strategies, and transformer-based architectures to solve lyric generation tasks. Qualitatively, the results of our MTurk study show that participants found our generated lyrics to be just as human-like and equally interesting as actual song lyrics suggesting that our line-level generation model is able to produce lyrics that are semantically meaningful and engaging, which are critical features to making successful pop music. Moreover interviews from the songwriters suggest new insights in lyricism, as a result of the models, that may not have been previously considered in developing their verses. 

We understand there exist limitations in our work, namely with the MTurk evaluation and usage of BLEU. Although all MTurk participants do not completely fit the audience for American pop music, given the presence of non-American participants, these members are still a valuable resource in understanding the effectiveness of the models. Another limitation lies in the brevity penalty of BLEU score calculations which uses length values averaged through the entire corpus. This leads to penalized scores for short lines like those in our training data that were shortened based on our syllable count pre-processing. Performing additional analysis using metrics like METEOR might allow us to better assess the quality of our lyrics.

Future work in lyric generation using transformer-based approaches could be the addition of more sophisticated rhyme schemes or the inclusion of context variables like mood, tone, style as other tasks for our models to learn in addition to the ones discussed in this paper. Alternative areas to consider include a transfer-learning based approach in combination with tagging pre-processing strategies for lyric generation in other musical genres or for different creative tasks such as story generation. Because of the increased BLEU score of our Genre Combined List model (compared to the regular Billboard data), we suspect that additional work with creating a more consistent dataset of pop lyrics could result in an increase in performance by matching the pop genre more closely. The proposed approach to solving this issue is to perform a much larger scale survey asking participants to evaluate which artists they consider purely ``pop" to rectify the subjectivity in the Genre dataset. Furthermore, having an additional MTurk study for songwriters and other domain experts might allow us to gather more insight and make stronger conclusions about the quality and ranking of our models. We anticipate our architecture to generalizable to other genres given pop music's history and influences.

``Say What?" aims to facilitate meaningful and novel interactions between humans and AI and to create novel outcomes through the collaborations between them. We hope to use this tool to inspire songwriters through these experiences and hope to keep making improvements as we continue to dialogue with and receive feedback on our model.

%\footnote{https://colab.research.google.com/drive/1zithQxrSSDgoabeNIF5iNj0dstjwcL6K?usp=sharing}
%\footnote{https://tinyurl.com/yywsy456}

%\appendix{\LaTeX{} and Word Style Files}\label{stylefiles}

%The \LaTeX{} and Word style files are available on the ICCC-13
%website, {\tt http://computationalcreativity.net/iccc2013/}.
%These style files implement the formatting instructions in this
%document.

%The \LaTeX{} files are {\tt iccc.sty} and {\tt iccc.tex}, and
%the Bib\TeX{} files are {\tt iccc.bst} and {\tt iccc.bib}. The
%\LaTeX{} style file is for version 2e of \LaTeX{}, and the Bib\TeX{}
%style file is for version 0.99c of Bib\TeX{} ({\em not} version
%0.98i).

%The Microsoft Word style file consists of a single template file, {\tt
%iccc.dot}. 

%These Microsoft Word and \LaTeX{} files contain the source of the
%present document and may serve as a formatting sample.  

\bibliographystyle{ACM-Reference-Format}
\bibliography{refs}

\end{document}